\documentclass[10pt,twocolumn,letterpaper]{article}

\usepackage[pagenumbers]{cvpr} % To force page numbers, e.g. for an arXiv version

\definecolor{cvprblue}{rgb}{0.21,0.49,0.74}
\usepackage[pagebackref,breaklinks,colorlinks,allcolors=cvprblue]{hyperref}

\usepackage{times}
\usepackage{epsfig}
\usepackage{graphicx}
\usepackage{amsmath}
\usepackage{amssymb}
\usepackage{bm}
\usepackage{float}
\usepackage{tabularx}
\usepackage{microtype}

\usepackage{booktabs} % for professional tables
\usepackage{url}
\usepackage{caption}
\usepackage{multirow}
\usepackage{multicol}
\usepackage{array}
\usepackage{pifont}
\usepackage{xcolor}
\usepackage{booktabs}
\usepackage[normalem]{ulem}
\usepackage{xspace}
\usepackage{diagbox}
\usepackage{color}
\usepackage{enumitem}
\usepackage{bm}
\usepackage{algorithm}
\usepackage{listings}
\usepackage{xspace}
\usepackage{makecell}
\usepackage{algorithm}
\usepackage{algpseudocode}

\newcommand{\method}{DMC}

\title{Preserving Cross-Modal Consistency for CLIP-based Class-Incremental Learning}

\author{%
  Haoran Chen$^{1,2}$~\hspace{5pt}
  Houze Xu$^{1,2}$~\hspace{5pt} 
  Micah Goldblum$^{3}$~\hspace{5pt} 
  Daoguo Dong$^{1,2}$~\hspace{5pt} 
  Zuxuan Wu$^{1,2}$~\hspace{5pt}
\vspace{0.1in}\\ 
$^1$Institute of Trustworthy Embodied AI, Fudan University \\
$^2$Shanghai Collaborative Innovation Center of Intelligent Visual Computing \\
$^{3}$Columbia University
}

\begin{document}
\maketitle
\begin{abstract}

Class-incremental learning (CIL) enables models to continuously learn new categories from sequential tasks without forgetting previously acquired knowledge. While recent advances in vision–language models such as CLIP have demonstrated strong generalization across domains, extending them to continual settings remains challenging. In particular, learning task-specific soft prompts for newly introduced classes often leads to severe classifier bias, as the text prototypes overfit to recent categories when prior data are unavailable. In this paper, we propose \method, a simple yet effective two-stage framework for CLIP-based CIL that decouples the adaptation of the vision encoder and the optimization of textual soft prompts. Each stage is trained with the other frozen, allowing one modality to act as a stable semantic anchor for the other to preserve cross-modal alignment. Furthermore, current CLIP-based CIL approaches typically store class-wise Gaussian statistics for generative replay, yet they overlook the distributional drift that arises when the vision encoder is updated over time. To address this issue, we introduce \method-OT, an enhanced version of \method~that incorporates an optimal-transport guided calibration strategy to align memory statistics across evolving encoders, along with a task-specific prompting design that enhances inter-task separability. Extensive experiments on CIFAR-100, Imagenet-R, CUB-200, and UCF-101 demonstrate that both \method~and \method-OT achieve state-of-the-art performance, with \method-OT further improving accuracy by an average of 1.80\%.

\end{abstract}    
\section{Introduction}
\label{sec:intro}

Class-incremental learning (CIL) enables models to continuously learn new classes from sequentially arriving tasks while preserving performance on previously learned ones~\cite{de2021defysurvey, masana2022classsurvey, wang2024comprehensivesurvey}. Unlike traditional training paradigms that assume joint access to all data, CIL reflects realistic scenarios where data and categories emerge over time. A central challenge in CIL is mitigating catastrophic forgetting~\cite{french1999catastrophic, kemker2018measuring, kirkpatrick2017overcoming}, where neural networks tend to overwrite prior knowledge when learning new classes, while maintaining a unified classifier capable of recognizing all classes seen so far.

\begin{figure}
    \centering
    \includegraphics[width=\linewidth]{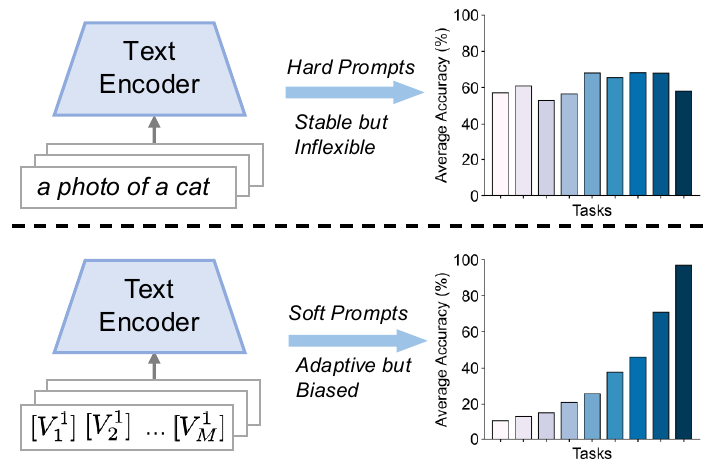}
    \caption{Comparison between hard and soft prompts in CLIP-based class-incremental learning. Handcrafted hard prompts (top) offer stable but inflexible representations, leading to consistent yet limited adaptation. In contrast, soft prompts (bottom) are adaptive but biased toward newly introduced classes, causing severe performance degradation for earlier tasks.}
    \label{fig:teaser}
\end{figure}

Recently, large pretrained vision–language models such as CLIP~\cite{radford2021learning} have revolutionized visual recognition by enabling powerful zero-shot generalization across diverse domains~\cite{jia2021scaling}. Naturally, this has sparked a growing interest in adapting CLIP to a wide range of downstream tasks~\cite{zhou2022extract, rao2022denseclip, weng2023open}. Among the various adaptation strategies, a major line of work builds upon replacing handcrafted hard prompts with learnable soft textual tokens to achieve richer, task-specific conditioning~\cite{lester2021power, zhou2022learning}. 

However, extending soft prompt learning to continual learning remains challenging. Under the CIL scenario, training task-specific soft prompts for newly introduced classes inevitably induces classifier bias~\cite{chen2024adaptive, wu2019large}: since each prompt essentially represents a class prototype, optimizing them without access to previous task data leads the classifier to overfit to recently seen categories.
As a result, most existing CLIP-based CIL works leverage manually crafted hard prompts and apply parameter-efficient finetuning strategies such as Adapters~\cite{houlsby2019parameter} or LoRA~\cite{hu2022lora} to directly modify the text encoder parameters~\cite{zhou2025learning, zhou2025external}.

In this work, we revisit soft prompt tuning from a new perspective and argue that its limitations primarily stem from the entangled optimization between visual and textual modalities. We hypothesize that by decoupling the training of the vision and text branches, the pretrained vision–language alignment of CLIP can be better preserved and exploited for continual adaptation. To this end, we propose a simple yet principled two-stage progressive adaptation framework. In the first stage, we adapt the vision encoder using handcrafted hard prompts while freezing the text encoder. This preserves the semantic anchor of CLIP’s textual space and ensures that the visual encoder remains aligned with the language domain during adaptation. In the second stage, we freeze the adapted vision encoder and, analogous to the frozen textual encoder in the first phase, use it as a stable anchor to optimize the textual soft prompts. Such a decoupled design enables progressive adaptation of both modalities while maintaining cross-modal consistency and mitigating classifier bias.

Following prior works~\cite{zhang2023slca, huang2024class, zhou2025external, chen2024promptfusion}, we also model each class’s visual features as a Gaussian distribution and store their means and covariances in memory. During subsequent tasks, synthetic features are generated by sampling from these stored Gaussians for generative replay~\cite{shin2017continual}, thereby helping to rebalance the textual classifier derived from soft prompts. We refer to this approach as \method~(\textbf{D}ual \textbf{M}odality Anchored \textbf{C}LIP-based CIL), and experimental results show that this minimal design can already achieve strong results across multiple benchmarks.

Nevertheless, a key limitation remains in this approach: once the vision encoder is updated, the previously stored Gaussian statistics no longer correspond to the true feature distributions of earlier classes, resulting in feature drift~\cite{theofilou2025stable, kapturowski2018recurrent} and degraded replay fidelity~\cite{caccia2021new}. In contrast, keeping the vision encoder frozen preserves stable feature statistics but prevents the model from adapting to the evolving data distribution. This dilemma raises a natural question: can we transform the old Gaussian statistics to their corresponding distributions under the updated encoder, using only data from the current task?

To address this, we further introduce an Optimal Transport (OT) guided calibration strategy~\cite{peyre2019computational, hoyos2019local}, \method-OT, to align class statistics across encoder updates. The goal is to learn an affine transformation that transports the mean and covariance of the old feature space to those of the updated encoder. Specifically, before training on the current task, we extract per-class features to compute their means and covariances. After adaptation, we re-extract features on the same data to obtain updated statistics, forming paired Gaussian representations before and after training. We then estimate a symmetric positive-definite affine transformation via the Wasserstein Optimal Transport map~\cite{korotin2021neural} to align these paired distributions. The learned mapping is subsequently applied to calibrate all previously stored Gaussian statistics, preserving consistency across evolving feature spaces. Meanwhile, we introduce a task-specific prompt design tailored for CIL, which explicitly enhances inter-task separability among the learned soft prompts.

We conduct extensive experiments on four widely used CIL benchmarks, namely CIFAR-100, Imagenet-R, CUB-200, and UCF-101, to validate the effectiveness of the proposed approach. Results show that \method~can already achieve state-of-the-art methods across all benchmarks, while the enhanced version, \method-OT, consistently achieves additional gains. On average, \method-OT improves over \method~by a significant margin of 1.80\%, confirming the benefit of our optimal transport calibration and task-specific prompting design. In summary, our main contributions are three-fold:
\begin{itemize}
\item We introduce \method, a two-stage training framework for CLIP-based CIL that decouples the adaptation of the vision encoder and the optimization of textual soft prompts. Each stage is trained with the other frozen, allowing one modality to serve as a stable anchor that preserves cross-modal alignment during adaptation.
\item We further introduce an enhanced variant, \method-OT, which incorporates an optimal-transport–guided calibration strategy to align Gaussian statistics across encoder updates, together with a task-specific prompting design that strengthens inter-task separability and enhances continual adaptation.
\item We perform comprehensive experiments on four benchmarks, demonstrating that both of our approaches achieve consistent and substantial improvements over previous state-of-the-art methods.
\end{itemize}
\section{Related Works}
\label{sec:related works}

\subsection{Vision-Language Models}

Large-scale vision–language pretraining has enabled unified visual–textual representations through contrastive learning. Among them, CLIP~\cite{radford2021learning} aligns image and text embeddings using hundreds of millions of web data, achieving remarkable zero-shot recognition. Its success inspired numerous variants, such as Align~\cite{jia2021scaling}, Filip~\cite{yao2021filip}, and SigLIP~\cite{zhai2023sigmoid}, which improved alignment granularity or training efficiency.

To adapt CLIP to downstream tasks, prompt-based and adapter-based~\cite{gao2024clip} tuning have become popular. CoOp~\cite{li2017learning} and CoCoOp~\cite{zhou2022conditional} learn soft textual prompts to specialize CLIP without updating backbones, while extensions like MaPLe~\cite{khattak2023maple} enhance generalization via visual prompts or semantic constraints. 

\subsection{Continual Learning}

Continual learning aims to enable models to acquire new knowledge over time while preserving previously learned information~\cite{de2021defysurvey}. A central challenge is catastrophic forgetting~\cite{kirkpatrick2017overcoming}, where new updates overwrite old representations. Early approaches address this issue through rehearsal~\cite{rebuffi2017icarl}, which stores exemplars from past tasks, or through regularization, which constrains parameter drift to protect important weights. More recent advances leverage pretrained backbones and parameter-efficient tuning strategies to alleviate forgetting without requiring explicit memory storage~\cite{chen2024adaptive}. For example, prompt-based methods such as L2P~\cite{wang2022l2p} and DualPrompt~\cite{wang2022dualprompt} dynamically allocate prompts for each task, while approaches like Coda-Prompt~\cite{smith2023coda} enhance retention via compositional or orthogonal prompt designs.

\subsection{CLIP-based Continual Learning}
Recently, CLIP has been extended to continual~\cite{zhou2025learning} and domain-adaptive settings~\cite{chen2023multi}, where maintaining vision–language alignment and mitigating forgetting become critical challenges. For instance, PROOF~\cite{zhou2025learning} introduces task-specific projection heads to fuse visual and textual representations while preserving previously learned knowledge, and CLAP4CLIP~\cite{jha2024clap4clip} employs probabilistic modeling over visual-guided text features to capture uncertainty and stabilize continual adaptation. With the growing trend of restricting the use of rehearsal data~\cite{wang2022l2p, wang2022dualprompt}, recent methods increasingly rely on generative replay to alleviate classifier bias induced by task-specific soft prompt tuning~\cite{zhang2023slca, chen2024promptfusion, zhou2025external}. However, the generative replay itself is constrained by the static nature of stored Gaussian statistics. When the vision encoder is trained, these distributions become misaligned with the updated feature space. This intrinsic limitation ultimately motivates us to propose \method-OT, which learns to transport feature memories across evolving encoders using optimal transport.

\begin{figure*}
    \centering
    \includegraphics[width=\linewidth]{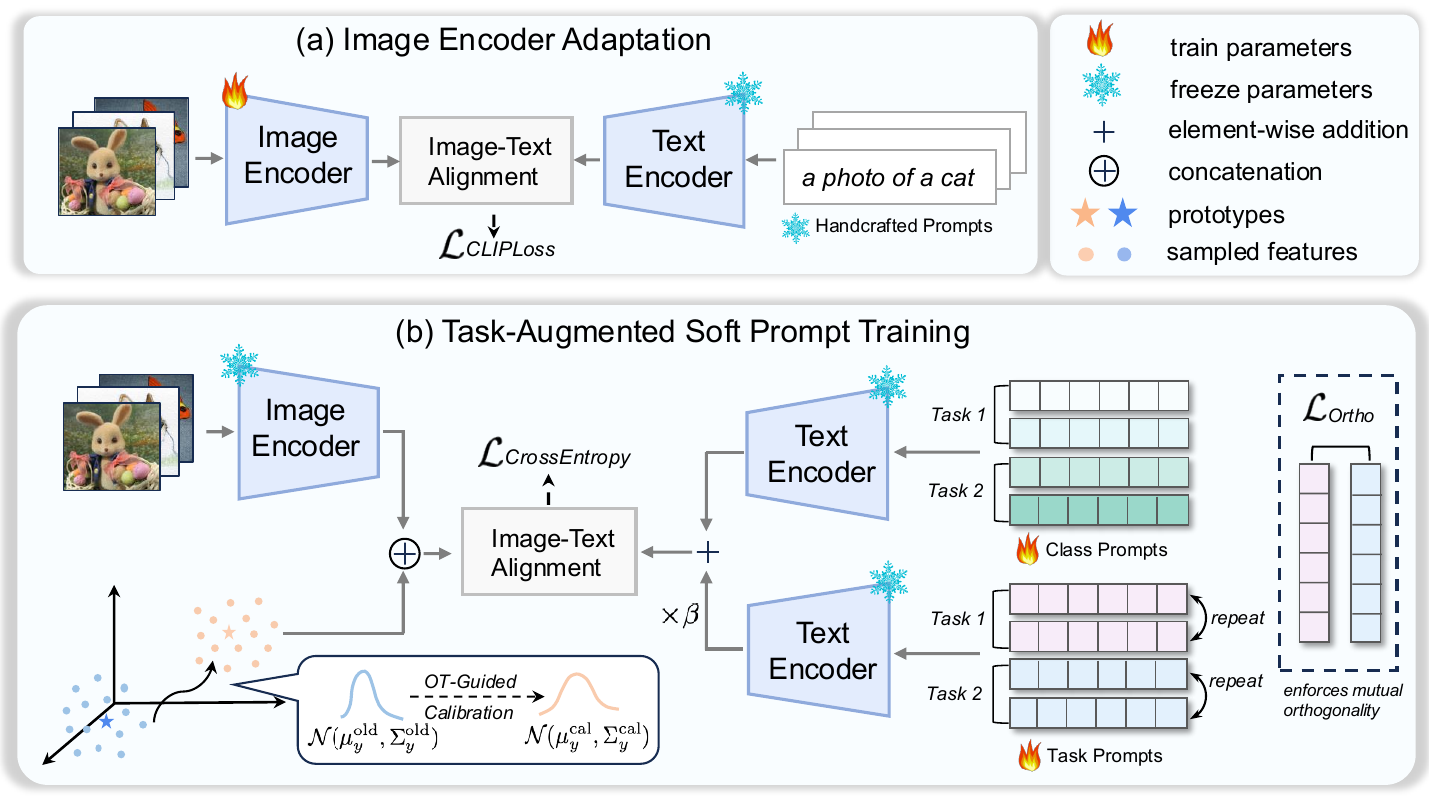}
    \caption{Overview of the proposed \method-OT training framework. (a) Image Encoder Adaptation: The image encoder is fine-tuned while freezing the text encoder with handcrafted hard prompts. This stage strengthens visual representations through the contrastive loss $\mathcal{L}_{\text{CLIP}}$. (b) Task-Augmented Soft Prompt Training: With the image encoder frozen, we train soft textual prompts via the cross-entropy loss $\mathcal{L}_{\text{CE}}$, combining class-specific and task-specific prompts through element-wise addition. The task-specific prompts are regularized by an orthogonality loss $\mathcal{L}_{\text{Ortho}}$ to ensure inter-task separability and feature diversity. During this stage, old calibrated class prototypes are modeled as multivariate Gaussians, from which features are sampled to implicitly replay past knowledge. }
    \label{fig:method}
\end{figure*}

\section{Method}

The goal of this paper is to develop a CLIP-based CIL framework focusing on soft prompt training. To this end, we first propose a simple yet principled two-stage framework that alternately adapts one modality while keeping the other fixed as a semantic anchor. Building upon this foundation, we further integrate an optimal-transport guided memory calibration mechanism together with a task-augmented prompt design to enhance feature consistency and inter-task separability. In the following sections, we first introduce the problem formulation and CLIP architecture in \cref{sec:preliminary}, then present the basic two-stage framework in \cref{sec:dmc}, followed by the optimal transport–based calibration strategy and task-augmented prompt design in \cref{sec:ot calibration} and \cref{sec:task prompt}, respectively. Finally, we summarize the overall training pipeline of the proposed \method-OT in \cref{sec:pipeline}. An overview of our method is shown in \cref{fig:method}.

\subsection{Preliminary}
\label{sec:preliminary}

Our work is conducted under the class-incremental learning setting using CLIP, where the model is trained sequentially on a series of tasks. In each task, a new set of classes is introduced, and the training data from previous tasks are strictly inaccessible, ensuring no overlap between old and new task data. Formally, we consider training the CLIP model across $K$ tasks, $\mathcal{T} = (T_1, T_2, ..., T_K)$, where each task is associated with a dataset $\mathcal{D}(\mathcal{X}_k, \mathcal{Y}_k)$. Here $\mathcal{Y}^i \cap \mathcal{Y}^j = \emptyset$ for $i \neq j$.

For the CLIP model, it consists of a vision encoder $f_v(\cdot)$ and a text encoder $f_t(\cdot)$, which projects an image $x$ and a text description $t$ into a shared embedding space:
\[
z_v = f_v(x), \quad z_t = f_t(t),
\]
where both embeddings are normalized, $\|z_v\|_2 = \|z_t\|_2 = 1$. At inference, the textual embedding of each class $y$ is computed by encoding a handcrafted prompt, most commonly in the form of ``a photo of a [CLS]'', as
\[
w_y = f_t(t_y),
\]
and the prediction for an image $x$ is obtained via cosine similarity between the image and all class text embeddings:
\[
p(y|x) = \frac{\exp(\text{sim}(f_v(x), w_y)/\tau)}{\sum_{y'\in \mathcal{Y}} \exp(\text{sim}(f_v(x), w_{y'})/\tau)}.
\]
In the CIL setting, as new classes are introduced incrementally, the textual classifier becomes increasingly biased toward recent classes, posing a central challenge that our method aims to address.

\subsection{Dual Modality Anchored Class-Incremental Learning}
\label{sec:dmc}

\paragraph{Image Encoder Adaptation} We begin by introducing the image encoder adaptation phase of \method. Given the $k$-th task with image set $\mathcal{X}_k$ and class set $\mathcal{Y}_k$, we update only the image encoder $f_v$, while keeping the text encoder $f_t$ frozen and using handcrafted hard prompts of the form “a photo of [CLS]” as textual inputs. This setup allows the model to leverage the rich semantic priors embedded in CLIP’s pretrained textual space as a stable anchor, ensuring that the adapted visual features remain well-aligned with the language domain during fine-tuning.

We adopt the original CLIP contrastive loss in this phase, which encourages matched image–text pairs to have similar embeddings while pushing apart unmatched ones. The rationale behind using this loss function is to ensure that the vision encoder continues to learn under the same pretraining paradigm as CLIP, thereby preserving the global vision–language alignment while facilitating smooth adaptation to the new task distribution. Formally, let $B$ denote the batch size, and for each sample $i$, let $z_v^i$ and $z_t^i$ represent the normalized image and text embeddings, respectively. With temperature $\tau$, the symmetric CLIP contrastive loss over the batch is defined as:

\begin{equation}
\label{eq:stage1-clip}
\begin{aligned}
\mathcal{L}_{\mathrm{CLIP}}
= -\frac{1}{2B}\sum_{i=1}^{B} \Bigg[
& \log
\frac{\exp(\mathrm{sim}(z_v^i, z_t^i)/\tau)}
     {\sum_{j=1}^{B}\exp(\mathrm{sim}(z_v^i, z_t^j)/\tau)} \\
& +\;
\log
\frac{\exp(\mathrm{sim}(z_t^i, z_v^i)/\tau)}
     {\sum_{j=1}^{B}\exp(\mathrm{sim}(z_t^i, z_v^j)/\tau)}
\Bigg].
\end{aligned}
\end{equation}

\paragraph{Soft Prompt Training} With the adapted vision encoder obtained from the first stage, we proceed to the second stage, which focuses on training soft textual prompts for continual adaptation. Specifically, for each class $y$ belonging to task $k$, we train class-specific prompt token parameters $P^{\text{cls}}_y \in \mathbb{R}^{M \times d}$, represented as ``$[V_1]...[V_M]$[CLS]”. Here, $M$ denotes the prompt length and $d$ is the dimensionality of CLIP’s embedding space.

Following common practice in CLIP-based CIL, we maintain class-wise feature statistics from previous tasks for implicit feature replay. Specifically, the model records the mean $\mu_y$ and covariance $\Sigma_y$ of the image features for each class $y$ and models the corresponding class distribution as a multivariate Gaussian $\mathcal{N}(\mu_y, \Sigma_y)$. When learning a new task, synthetic features sampled from these stored Gaussians are combined with the current task features for generative replay, effectively mitigating classifier bias of the learned prompts toward newly introduced classes.

Unlike the previous stage where the vision encoder is optimized via the contrastive objective, the textual prompts are trained using a standard cross-entropy loss over all classes encountered so far. This design choice follows empirical findings in prior works~\cite{zhou2022learning, lu2022prompt}, which show that direct classification supervision is effective for learning discriminative prompt representations. Formally, given an image $i$ with feature $z_v^i$ from the frozen image encoder and soft text embeddings $\{w_{y,k}^{\text{soft}}\}_{y \in \mathcal{Y}_{all}}$, the loss is computed as:
\begin{equation}
\label{eqn:ce_loss}
\mathcal{L}_{\text{CE}} = -\frac{1}{B}\sum_{i=1}^{B} 
\log \frac{\exp(\mathrm{sim}(z_v^i, w_{y_i,k}^{\text{soft}})/\tau)}
{\sum_{y \in \mathcal{Y}_{all}} \exp(\mathrm{sim}(z_v^i, w_{y,k}^{\text{soft}})/\tau)}.
\end{equation}
where $\mathcal{Y}_{all}$ denotes the cumulative set of all classes seen up to the current task, and $B$ is the batch size. 

\subsection{Optimal Transport Guided Calibration}
\label{sec:ot calibration}

While the above framework already demonstrates strong performance, several challenges remain. Most notably, when the CLIP vision encoder is updated sequentially, the previously stored Gaussian statistics for generative replay no longer correspond to the true feature distributions under the updated encoder. To address this, we introduce an Optimal Transport guided calibration strategy that aligns class statistics across encoder updates. 

Concretely, before training on the current task, we extract per-class features and compute their mean and covariance $(\mu_y^{\text{pre}}, \Sigma_y^{\text{pre}})$. After adapting the vision encoder on the current task, we re-extract features on the same data and obtain their updated statistics $(\mu_y^{\text{post}}, \Sigma_y^{\text{post}})$. We then average the class statistics:
\[
\begin{aligned}
    \bar{\mu}^{\text{pre}} = \frac{1}{|\mathcal{Y}_k|}\sum_{y\in\mathcal{Y}_k}\mu_y^{\text{pre}}, \quad
    \bar{\Sigma}^{\text{pre}} = \frac{1}{|\mathcal{Y}_k|}\sum_{y\in\mathcal{Y}_k}\Sigma_y^{\text{pre}} \\
       \bar{\mu}^{\text{post}} = \frac{1}{|\mathcal{Y}_k|}\sum_{y\in\mathcal{Y}_k}\mu_y^{\text{post}}, \quad
    \bar{\Sigma}^{\text{post}} = \frac{1}{|\mathcal{Y}_k|}\sum_{y\in\mathcal{Y}_k}\Sigma_y^{\text{post}}
\end{aligned}
\]
These averaged statistics form a global Gaussian correspondence that reflects the encoder’s feature-space transformation on the current task:
\[
    \mathcal{N}(\bar{\mu}^{\text{pre}},\bar{\Sigma}^{\text{pre}}) \rightarrow \mathcal{N}(\bar{\mu}^{\text{post}},\bar{\Sigma}^{\text{post}})
\]

We then compute the Wasserstein Optimal Transport map~\cite{korotin2021neural, peyre2019computational} between these two sets of distributions. Formally, given two Gaussian distributions $\mathcal{N}(\bar{\mu}^{\text{pre}},\bar{\Sigma}^{\text{pre}})$ and $\mathcal{N}(\bar{\mu}^{\text{post}},\bar{\Sigma}^{\text{post}})$, the 2-Wasserstein distance between them has a closed-form solution:
\begin{equation}
\begin{aligned}
    W_2^2 = &\|\bar{\mu}^{\text{pre}} - \bar{\mu}^{\text{post}}\|_2^2 
   \quad + \quad \\
    &\mathrm{Tr}(\bar{\Sigma}^{\text{pre}} + \bar{\Sigma}^{\text{post}} 
    - 2\big((\bar{\Sigma}^{\text{pre}})^{1/2}
        \bar{\Sigma}^{\text{post}}
        (\bar{\Sigma}^{\text{pre}})^{1/2}\big)^{1/2}).
\end{aligned}
\end{equation}
The corresponding Optimal Transport map that minimizes this distance is an affine transformation:
\begin{equation}
\label{eqn:ot map}
\begin{aligned}
    \mathbf{T} &= (\bar{\Sigma}^{\text{pre}})^{-1/2}((\bar{\Sigma}^{\text{pre}})^{1/2}\bar{\Sigma}^{\text{post}}(\bar{\Sigma}^{\text{pre}})^{1/2})^{1/2}(\bar{\Sigma}^{\text{pre}})^{-1/2} \\ 
    \mathbf{b} &= \bar{\mu}^{\text{post}} - \mathbf{T}\bar{\mu}^{\text{pre}},
\end{aligned}
\end{equation}
Intuitively, $\mathbf{T}$ captures how the geometry of the feature space deforms as the encoder evolves, while $\mathbf{b}$ aligns the mean displacement between feature distributions. During computation, we employ the Ledoit–Wolf shrinkage~\cite{ledoit2004well} estimator to regularize the covariance matrix and improve its numerical stability, thereby facilitating reliable matrix inversion.

We apply this learned transformation to all previously stored Gaussian statistics, obtaining calibrated parameters:
\[
    \mu_y^{\text{cal}} = \mathbf{T} \mu_y^{\text{old}} + \mathbf{b}, \quad
    \Sigma_y^{\text{cal}} = \mathbf{T} \Sigma_y^{\text{old}} \mathbf{T}^\top.
\]
The calibrated Gaussians $\mathcal{N}(\mu_y^{\text{cal}}, \Sigma_y^{\text{cal}})$ are then used for generative replay in the soft prompt training stage, ensuring that replayed features remain consistent with the continually evolving encoder representation.

\begin{algorithm}[t]
\caption{Training Pipeline for \method-OT}
\label{alg:pipeline}
\begin{algorithmic}[1]
\Require CLIP model $(f_v, f_t)$; task data $\mathcal{D}_k=(\mathcal{X}^k,\mathcal{Y}^k)$; stored stats $\{(\mu_y^{\text{old}},\Sigma_y^{\text{old}})\}_{y\in\mathcal{Y}_{1:k-1}}$
\vspace{1ex}
\Statex \textbf{Before Training}
\State Compute per-class features with current $f_v$: $z_v \leftarrow f_v(\mathcal{X}_k)$ and class stats $\{(\mu_y^{\text{pre}},\Sigma_y^{\text{pre}})\}_{y\in\mathcal{Y}_k}$ 
\vspace{1ex}
\Statex \textbf{Image Encoder Adaptation} 
\State Train $f_v$ on $\mathcal{D}_k$ using hard prompts \Comment{\cref{eq:stage1-clip}}
\vspace{1ex}
\Statex \textbf{OT Calibration}
\State Recompute per-class features and class stats $\{(\mu_y^{\text{post}},\Sigma_y^{\text{post}})\}_{y\in\mathcal{Y}_k}$ with updated $f'_v$
\State Compute $(\mathbf{T},\mathbf{b})$ \Comment{\cref{eqn:ot map}}
\State Calibrate all previously stored stats:
\Statex \hspace{1.1em}$\mu_y^{\text{cal}} \!\leftarrow\! \mathbf{T}\mu_y^{\text{old}} + \mathbf{b}, \quad
\Sigma_y^{\text{cal}} \!\leftarrow\! \mathbf{T}\Sigma_y^{\text{old}}\mathbf{T}^{\top} \ \ \forall y\!\in\!\mathcal{Y}_{1:k-1}$
\vspace{1.5ex}
\Statex \textbf{Task-Augmented Soft Prompt Training}
\State Initialize class prompts $\{P_y^{\text{cls}}\}_{y\in\mathcal{Y}_k}$ and task prompt $P_k^{\text{tsk}}$. Form soft text embeddings \Comment{\cref{eqn:soft embedding}}
\State Sample synthetic feature from $\{\mathcal{N}(\mu_y^{\text{cal}}, \Sigma_y^{\text{cal}})\}_{y\in\mathcal{Y}_{1:k-1}}$
\State Train textual prompts on $\mathcal{D}_k$ and synthetic data 
\Statex \Comment{\cref{eqn:total_loss}}
\State Store new stats $\{(\mu_y,\Sigma_y)\}_{y\in\mathcal{Y}_k}$ for the next task
\end{algorithmic}
\end{algorithm}
\vspace{2ex}

\subsection{Task-Augmented Prompt Design}
\label{sec:task prompt}

The prompt design introduced in the previous section primarily targets adaptation in static environments. However, such a design often lacks sufficient inter-task discrimination under the CIL setting. To enhance both feature separability and cross-task adaptability, we introduce a set of task-specific prompts $P^{\text{tsk}}_k \in \mathbb{R}^{M \times d}$. Specifically, for each task comprising $N_k$ classes, in addition to learning $N_k$ class-specific prompts, we also train \textit{one} task-specific prompt shared across all classes within the task. This results in a total of $K$ task prompts for $K$ tasks. The final text embedding for class $y$ from task $k$ is obtained by element-wise combining the encoded class and task prompts as follows:
\begin{equation}
\label{eqn:soft embedding}
w_{y,k}^{\text{soft}} = 
\frac{f_t(P_y^{\text{cls}}) + \beta \cdot f_t(P_k^{\text{tsk}})}
{\big\| f_t(P_y^{\text{cls}}) + \beta \cdot f_t(P_k^{\text{tsk}}) \big\|_2},
\end{equation}
% \vspace{1ex}
where \( \beta \) is a scaling factor that balances the relative contribution of the task-specific prompt to the overall representation.

The goal of introducing task-specific prompts is to encourage richer task diversity. To achieve this, we apply an \emph{orthogonality regularization} to their feature representations. This helps promoting pairwise orthogonality and thereby improving inter-task separability:
\begin{equation}
\label{eqn:ortho_loss}
\mathcal{L}_{\text{ortho}}
= \frac{1}{K(K-1)} \sum_{\substack{i\neq j}} 
\big\| (f_t(P^{\text{tsk}}_i))^\top f_t(P^{\text{tsk}}_j) \big\|_F^2.
\end{equation}
Bringing this with \cref{eqn:ortho_loss}, the objective for training the textual prompts after task augmentation is:
\begin{equation}
\label{eqn:total_loss}
\mathcal{L} \;=\; \mathcal{L}_{\text{CE}} \;+\; \lambda_{\text{ortho}} \, \mathcal{L}_{\text{ortho}},
\end{equation}
with $\lambda_{\text{ortho}}$ controlling the strength of the regularizer.

\subsection{Overall Training Pipeline for \method-OT}
\label{sec:pipeline}

We summarize in \cref{alg:pipeline} the overall training process of \method-OT. Given the pretrained CLIP model and previously stored class statistics, we first adapt the vision encoder using a CLIP-based contrastive objective to capture task-specific visual patterns. To prevent distributional drift caused by sequential encoder updates, we then apply the Optimal Transport guided calibration to align historical Gaussian statistics with the new feature space. Finally, we perform task-augmented soft prompt training using both real and calibrated synthetic features to refine the textual representations and reinforce inter-task discrimination.

\section{Experiments}
\label{sec:experiments}

\subsection{Experimental Setup}
\paragraph{Datasets} We evaluate our method on four widely used benchmarks: CIFAR-100~\cite{krizhevsky2009cifar}, Imagenet-R~\cite{hendrycks2021imgnetr}, CUB-200~\cite{wah2011cub}, and UCF-101~\cite{soomro2012ucf101}. CIFAR-100 is a standard image classification dataset comprising 60,000 images of size 32 $\times$ 32 pixels. Imagenet-R is a challenging benchmark designed to evaluate model robustness to natural distribution shifts. It contains 30,000 images across 200 Imagenet classes, featuring diverse artistic renditions such as cartoons, graffiti, origami, and more. The combination of both semantic and covariate shifts makes Imagenet-R one of the most difficult datasets for continual learning. CUB-200, on the other hand, is a fine-grained visual classification dataset consisting of 11,788 images spanning 200 bird species. Owing to strong intra-class variation and visually similar inter-class samples, it serve as an effective benchmark for evaluating continual adaptation. Finally, UCF-101 is a widely used human-action recognition dataset containing 13,320 video clips across 101 action categories. Following prior work~\cite{zhou2025learning}, we sample 100 classes from it for training and evaluation setup.

\noindent\textbf{Evaluation metrics.} In this work, we adopt two widely used evaluation metrics for class-incremental learning of CLIP. The first one is $\mathcal{A}_B$ that measures the performance of the model after the last training session. The second one is $\bar{\mathcal{A}}$, which is the averaged performance along incremental stages. Formally, let $R_{k,i}$ denote the classification accuracy on task $T_i$ after training on task $T_k$. 
Then:
\[
\mathcal{A}_B = \frac{1}{K}\sum_{i=1}^{K} R_{K,i}, \quad \quad \bar{\mathcal{A}} = \frac{1}{B}\sum_{b=1}^{B}\mathcal{A}_b.
\]

\subsection{Implementation Details}
% All experiments are conducted on a single NVIDIA RTX 4090 GPU. 
Following prior works~\cite{huang2024class, chen2024promptfusion}, we adopt the ViT-B/16 architecture of CLIP pretrained by OpenAI as our backbone. The AdamW optimizer is used with a batch size of 32 across all datasets and both training stages. The textual prompt length $M$ is fixed at 10 for all datasets. The scaling coefficient $\beta$ is set to 0.1 for ImageNet-R and UCF-101, and to 0.05 and 1.5 for CIFAR-100 and CUB-200, respectively. The orthogonality regularization weight $\lambda_{\text{ortho}}$ in Eqn.~\ref{eqn:total_loss} is set to 0.5 for CUB-200 and 0.1 for the remaining datasets.

For the main experiments, we compare our method with recent state-of-the-art approaches, including Zero-Shot CLIP~\cite{radford2021learning}, CoOp~\cite{zhou2022learning}, SimpleCIL~\cite{zhou2024continual}, L2P~\cite{wang2022l2p}, DualPrompt~\cite{wang2022dualprompt}, Coda-Prompt~\cite{smith2023coda}, PROOF~\cite{zhou2025learning}, CLAP4CLIP~\cite{jha2024clap4clip}, RAPF~\cite{huang2024class}, and ENGINE~\cite{zhou2025external}. For fair comparison, all methods are re-implemented using the same CLIP backbone and task order. Specifically, SimpleCIL, L2P, DualPrompt, and Coda-Prompt are vision-only methods, which we implement using CLIP’s visual encoder. The remaining methods are CLIP-based. Among them, CoOp, PROOF, and CLAP4CLIP requires external memory, whereas the others do not.

\begin{table*}[!t]
\begin{center}
\setlength{\tabcolsep}{5mm}
\resizebox{\linewidth}{!}{
\begin{tabular}{l|c|cccccccccc}
\toprule 
\multirow{2}{*}{\textbf{Method}} & \multirow{2}{*}{\textbf{Memory}} &\multicolumn{2}{c}{\textbf{CIFAR-100}}
 &  \multicolumn{2}{c}{\textbf{Imagenet-R}}&  \multicolumn{2}{c}{\textbf{CUB-200}} &  \multicolumn{2}{c}{\textbf{UCF-101}}\\ 
& & \multicolumn{1}{c}{$\bar{\mathcal{A}}$}  &\multicolumn{1}{c}{$\mathcal{A}_B$}  & \multicolumn{1}{c}{$\bar{\mathcal{A}}$} & \multicolumn{1}{c}{$\mathcal{A}_B$}& \multicolumn{1}{c}{$\bar{\mathcal{A}}$} & \multicolumn{1}{c}{$\mathcal{A}_B$}  & \multicolumn{1}{c}{$\bar{\mathcal{A}}$} & \multicolumn{1}{c}{$\mathcal{A}_B$} \\
\midrule

ZS-CLIP~\cite{radford2021learning}  & \ding{55}& -  & 62.97& -  & 69.68& - & 47.54   & - & 65.21 \\

\midrule
CoOp~\cite{zhou2022learning}  & \ding{51} & 75.34 & 71.02 & 72.47 & 68.53 & 72.37 & 69.42 & 88.74 & 83.75\\
PROOF~\cite{zhou2025learning}  & \ding{51} & 83.92 & \underline{79.85} & 82.21 & 78.48 & 81.53 & 77.60 & 93.62 & 88.94\\
CLAP4CLIP~\cite{jha2024clap4clip} & \ding{51} & 83.43 & 79.12 & 84.04 & 79.63 & 80.55 & 76.32 & 91.35 & 86.42\\

\midrule
SimpleCIL~\cite{zhou2024continual} & \ding{55} & 74.52 & 68.73 & 75.17 & 68.05 & 81.35 & 73.55 & 89.73 & 84.58\\
L2P~\cite{wang2022l2p} & \ding{55} & 78.97 & 71.06 & 78.32 & 71.57 & 70.95 & 65.43 & 88.46 & 81.62\\
DualPrompt~\cite{wang2022dualprompt} & \ding{55} & 72.84 & 65.25 & 73.45 & 68.83 & 66.54 & 60.97 & 90.95 & 84.83\\
Coda-Prompt~\cite{smith2023coda} & \ding{55} & 74.46 & 69.82 & 71.96 & 65.36 & 68.12 & 62.78 & 87.73 & 82.25\\
RAPF~\cite{huang2024class}& \ding{55} & 84.07 & 79.25 & \underline{84.78} & 80.33 & \textbf{83.35} & 76.52 & 92.52 & 87.45\\
ENGINE~\cite{zhou2025external}& \ding{55} & 82.58 & 78.67 & 83.77 & 77.32 & 81.84 & 75.83 & \underline{95.04} & 90.11\\
 
\midrule
\method~(Ours) & \ding{55} & \underline{84.37} & 79.57  & 84.62 & \underline{81.55} & 82.76 & \underline{77.65} &94.80 & \underline{90.63} \\
\method-OT~(Ours) & \ding{55} & \textbf{86.13}& \textbf{82.16} & \textbf{85.05} & \textbf{83.01} & \underline{83.13} & \textbf{78.11} &\textbf{95.64} & \textbf{93.31}\\

\bottomrule
\end{tabular}}
\caption{Performance comparisons with CLIP-based CIL methods on CIFAR-100, Imagenet-R, CUB-200, and UCF-101 under a 10-task continual learning setting. Best results are marked in \textbf{bold}. Second best results are \underline{underscored}.} 
\vspace{-2ex}
\label{table:main_results}
\end{center}
\end{table*}

\subsection{Comparisons Against State-of-the-Art}

In this section, we compare our method against both memory-based and memory-free CLIP-based continual learning approaches across four benchmarks under a 10-task incremental setting, with results summarized in \cref{table:main_results}.

Among methods that utilize replay buffers, our approach achieves consistently superior performance across all datasets despite using no external memory. Specifically, \method~surpasses PROOF by an average of $1.13\%$ and CLAP4CLIP by $1.98\%$, demonstrating that our framework can effectively alleviate forgetting without explicit data replay. For the enhanced variant, \method-OT, the performance gain is even larger, exceeding $3\%$ on average across benchmarks. We also implement a memory-based version of CoOp, which was originally designed for prompt learning in static settings. Both \method~and \method-OT~still outperform this variant by a substantial margin, highlighting the strength of our framework does not lie in this specific choice of parameter-efficient finetuning.

When compared to memory-free state-of-the-art methods, our method continues to deliver notable improvements. On the widely used CIFAR-100 and Imagenet-R benchmarks, \method~outperforms the recent RAPF and ENGINE by an average of $0.77\%$ and $2.61\%$, respectively. With the inclusion of optimal transport calibration and task-augmented prompting design, \method-OT~further boosts performance, achieving gains of $3.20\%$ and $4.19\%$, respectively. On the fine-grained CUB-200 dataset, our method also attains the best $\mathcal{A}_B$ performance. Interestingly, the improvement of \method-OT~over \method~is relatively modest (0.37$\%$), a phenomenon we further investigate in the ablation study. Finally, on UCF-101, \method-OT~surpasses all existing approaches by a significant margin ($>3\%$), with \method~achieving the second-best overall result.

In summary, our approach establishes new state-of-the-art results across all four benchmarks, outperforming both memory-based and memory-free baselines. Notably, even the base version \method, which requires no additional modules or complex regularization, achieves strong performance through its simple yet principled design. These results validate the core advantages of our framework.

\begin{figure*}[t]
  \centering
   \includegraphics[width=0.85\linewidth]{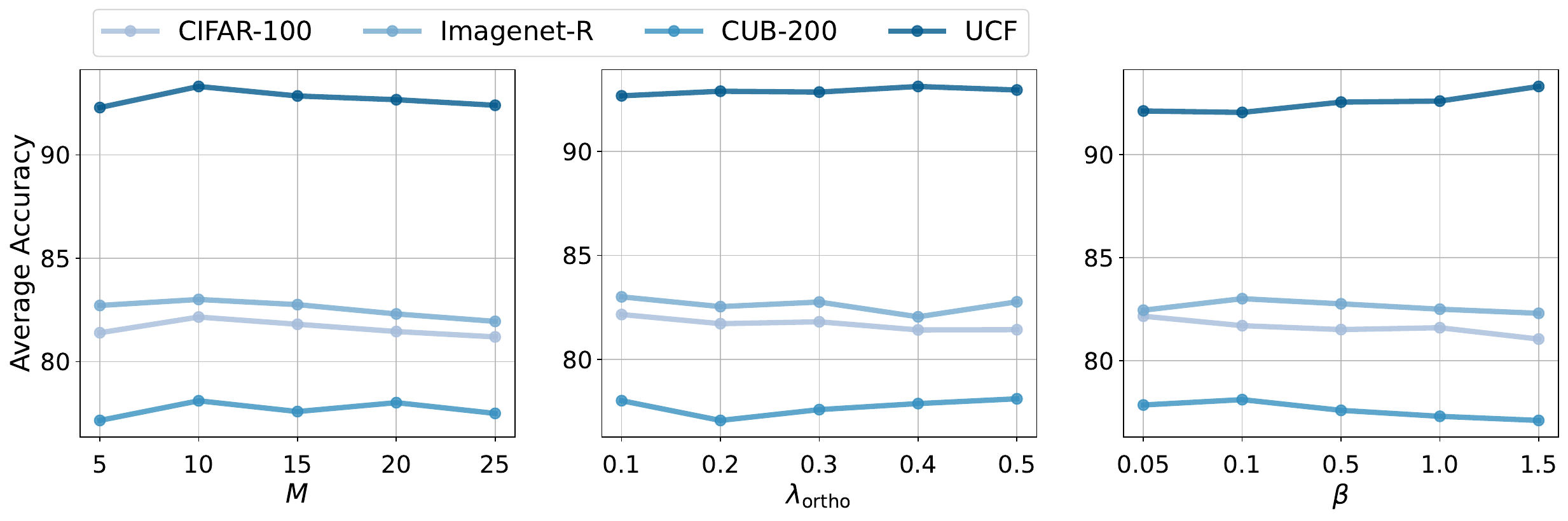}
      \vspace{-1ex}
   \caption{Hyperparameter sensitivity analysis.}
   \label{fig:hyperparameter}
   \vspace{-1ex}
\end{figure*}

\subsection{Long Task Sequence}

We further evaluate our framework under longer task horizons and conduct experiments on 20-task splits of CIFAR-100 and Imagenet-R, respectively. We report the results in \cref{tab:long task results}. Across both datasets, our method maintains strong performance and continues to outperform all baselines. On CIFAR-100, \method~already surpasses most state-of-the-art approaches. Its performance is only $0.33\%$ and $0.61\%$ less than PROOF and RAPF, respectively. The enhanced \method-OT, on the other hand, achieves the best results overall. A similar trend holds on Imagenet-R. \method~remains competitive performance with only $0.98\%$ lower than RAPF, and \method-OT~further boosts accuracy by a significant margin, achieving the highest performance among all methods. This demonstrates the robustness of both \method~and \method-OT.

\begin{table}[h]
\begin{center}
\setlength{\tabcolsep}{5mm}
\resizebox{\linewidth}{!}{
\begin{tabular}{l|ccccc}
\toprule 
\multirow{2}{*}{\textbf{Method}}  &\multicolumn{2}{c}{\textbf{CIFAR-100}}
 &  \multicolumn{2}{c}{\textbf{ImageNet-R}}\\ 
&\multicolumn{1}{c}{$\bar{\mathcal{A}}$}  &\multicolumn{1}{c}{$\mathcal{A}_B$}  & \multicolumn{1}{c}{$\bar{\mathcal{A}}$} & \multicolumn{1}{c}{$\mathcal{A}_B$} \\
\midrule

ZS-CLIP~\cite{radford2021learning} & -  & 62.97& -  & 69.68 \\

\midrule
CoOp~\cite{zhou2022learning}  & 73.42 & 69.37 & 70.11 & 65.70\\
PROOF~\cite{zhou2025learning} & 82.62 & 78.45 & 82.35 & 76.78 \\
CLAP4CLIP~\cite{jha2024clap4clip}  & 81.31 & 77.82 & 83.12 & 77.08 \\

\midrule
SimpleCIL~\cite{zhou2024continual}  & 72.12 & 65.03 & 73.80 & 66.23 \\
L2P~\cite{wang2022l2p}  & 74.63 & 66.82 & 75.02 & 68.31 \\
DualPromp~\cite{wang2022dualprompt}  & 69.31 & 62.55 & 70.45 & 65.78\\
Coda-Prompt~\cite{smith2023coda}  & 72.13 & 67.61 & 69.32 & 64.12\\
RAPF~\cite{huang2024class} & 83.27 & \underline{78.73} & \underline{83.33} & \underline{79.63} \\
ENGINE~\cite{zhou2025external} & 81.88 & 77.50 & 81.05 & 76.15 \\
 
\midrule
\method~(Ours)  & \underline{83.36} & 78.12  & 82.97 & 78.65 \\
\method-OT~(Ours) & \textbf{85.44}& \textbf{81.00} & \textbf{84.12} & \textbf{80.42} \\

\bottomrule
\end{tabular}}
\caption{Performance comparisons on CIFAR-100 and Imagenet-R under a 20-task continual learning setting.} 
\vspace{-3ex}
\label{tab:long task results}
\end{center}
\end{table}

\subsection{Ablation Study}

\paragraph{Simultaneous Training} We first study the impact of our decoupled two-stage training design. To this end, we train \method~under a simultaneous optimization setting, where both the vision encoder and the textual soft prompts are updated jointly within a single stage using the same training data. As shown in \cref{tab:joint train}, this joint optimization strategy leads to a clear performance drop across all benchmarks. Specifically, the average accuracy decreases by $2.91\%$ compared to the proposed two-stage framework. We attribute this degradation to the interference between modalities: updating both encoders concurrently disrupts CLIP’s pretrained vision–language alignment, causing the feature space and text classifier overfit to the current task distribution.

\begin{table}[h] 
% \vspace{-1ex}
\tabcolsep = 1mm
\centering
\resizebox{\linewidth}{!}{
\begin{tabular}{c|cccc}
\toprule
 &  CIFAR-100& Imagenet-R  & CUB-200 & UCF-101\\
 \midrule
 \midrule
Simultaneous Training     &76.47     &79.06     &75.91  & 86.35  \\
Two-Stage Training & \textbf{79.57} & \textbf{81.55} & \textbf{77.65} & \textbf{90.64}\\

\bottomrule
\end{tabular}}
\caption{Ablation studies on two-stage training design.}
\vspace{-2ex}
\label{tab:joint train}
\end{table}

\paragraph{Optimal Transport Mapping} We next assess the effectiveness of the proposed OT-based calibration. In particular, we remove the OT mapping and directly replay features from the uncalibrated Gaussian statistics. As shown in \cref{tab:ot ablation}, this leads to an average performance drop of $1.67\%$ across all datasets, indicating that storing feature statistics alone is sub-optimal once the encoder evolves.

We further ablate our design choice of computing a single affine transport map using the averaged mean and covariance of the current task. As an alternative, we compute a separate OT map for each class pair and then average the resulting transformation parameters. This variant yields consistently lower performance, and notably results in a substantial degradation on the fine-grained CUB-200 dataset. Given its limited overall improvement with OT ($0.16\%$), we hypothesize that OT-based calibration may face inherent challenges on fine-grained distributions, suggesting an interesting direction for future investigation.

\begin{table}[h] 
% \vspace{-1ex}
\tabcolsep = 1mm
\centering
\resizebox{0.9\linewidth}{!}{
\begin{tabular}{c|cccc}
\toprule
 &  CIFAR-100& Imagenet-R  & CUB-200 & UCF-101\\
 \midrule
 \midrule
 w/o OT   &79.37     &82.02 & 77.95 & 90.56\\
with alternative OT     &81.55     &82.78     &62.24  & 92.93  \\

\midrule
with OT & \textbf{82.16} & \textbf{83.01} & \textbf{78.11} & \textbf{93.31}\\

\bottomrule
\end{tabular}}
\caption{Ablation study on the use of Optimal Transport.}
\vspace{-3ex}
\label{tab:ot ablation}
\end{table}

\paragraph{Task-Augmented Prompt Design} Finally, we evaluate the effectiveness of the task-augmented prompt design. As shown in \cref{tab:prompt ablation}, removing this structure leads to an average drop of $0.75\%$ across all datasets. These results indicate that the task-augmented design successfully balances shared and task-specific information, yielding more robust representations for continual learning.

\begin{table}[h] 
% \vspace{-1ex}
\tabcolsep = 1mm
\centering
\resizebox{\linewidth}{!}{
\begin{tabular}{c|cccc}
\toprule
 &  CIFAR-100& Imagenet-R  & CUB-200 & UCF-101\\
 \midrule
 \midrule
 w/o task-augment   &81.81     &82.41 & 76.92 & 92.45\\
with task-augment & \textbf{82.16} & \textbf{83.01} & \textbf{78.11} & \textbf{93.31}\\

\bottomrule
\end{tabular}}
\caption{Ablation study on the use of task-augmented prompts.}
\vspace{-2ex}
\label{tab:prompt ablation}
\end{table}

\subsection{Hyperparameter Analysis}
We conduct a sensitivity analysis on the three key hyperparameters: prompt length $M$, scaling coefficient $\beta$, and orthogonality regularization weight $\lambda_{\text{ortho}}$ across all four datasets. The results are summarized in \cref{fig:hyperparameter}. Overall, the performance of \method-OT remains stable under a wide range of hyperparameter choices.

\section{Conclusion}

In this work, we proposed \method, a simple yet principled two-stage framework that alternately updates one modality while using the other as a semantic anchor for CLIP-based class-incremental learning. Building on this foundation, we further introduced \method-OT, which leverages optimal transport to calibrate stored Gaussian statistics with the evolving feature space, thereby achieving implicit replay without direct access to past data. Experiments showed that both methods achieve state-of-the-art performances.

Through this study, we hope our findings inspire deeper exploration into multimodal continual learning, where the interplay between vision and language presents not only a challenge but also a unique opportunity to develop models that learn persistently and adaptively over time. Looking ahead, extending the proposed design philosophy to large-scale multimodal systems such as MLLM and VLA frameworks represents an exciting direction for future research. Such extensions could bridge continual adaptation with the scalability and compositional reasoning capabilities of foundation models, ultimately paving the way toward more unified and lifelong multimodal intelligence.

% \newpage
{
    \small
    \bibliographystyle{ieeenat_fullname}
    \bibliography{main}
}

% WARNING: do not forget to delete the supplementary pages from your submission 
% \input{sec/X_suppl}

\end{document}